\documentclass[10pt, conference, compsocconf]{IEEEtran}
\ifCLASSINFOpdf
  \usepackage[pdftex]{graphicx}
\else
\fi
\usepackage[cmex10]{amsmath}
\usepackage{amssymb}

\usepackage{tikz}
\usepackage{textcomp}
\usepackage{lipsum}

\newcommand\copyrighttext{%
	\footnotesize \textcopyright 2021 IEEE. Personal use of this material is permitted.
	Permission from IEEE must be obtained for all other uses, in any current or future
	media, including reprinting/republishing this material for advertising or promotional
	purposes, creating new collective works, for resale or redistribution to servers or
	lists, or reuse of any copyrighted component of this work in other works.}
\newcommand\copyrightnotice{%
	\begin{tikzpicture}[remember picture,overlay]
	\node[anchor=south,yshift=10pt] at (current page.south) {\fbox{\parbox{\dimexpr\textwidth-\fboxsep-\fboxrule\relax}{\copyrighttext}}};
	\end{tikzpicture}%
}

\begin{document}
	
\bstctlcite{IEEEexample:BSTcontrol}
%
\title{ORSA: Outlier Robust Stacked Aggregation for Best- and Worst-Case Approximations of Ensemble Systems\\
}


\author{
	\IEEEauthorblockN{Peter Domanski, Dirk Pflüger}
	\IEEEauthorblockA{\textit{Institute for Parallel and Distributed Systems} \\
		\textit{University of Stuttgart}\\
		Stuttgart, Germany \\
		\{peter.domanski, dirk.pflueger\}@ipvs.uni-stuttgart.de} 
	\and
	\IEEEauthorblockN{Raphaël Latty, Jochen Rivoir}
	\IEEEauthorblockA{\textit{Advantest Europe GmbH} \\
		\textit{Applied Research and Venture Team}\\
		\{raphael.latty, jochen.rivoir\}@avantest.com}
}


%


\maketitle
\copyrightnotice

\begin{abstract}
In recent years, the usage of ensemble learning in applications has grown significantly due to increasing computational power allowing the training of large ensembles in reasonable time frames. Many applications, e.g., malware detection, face recognition, or financial decision-making, use a finite set of learning algorithms and do aggregate them in a way that a better predictive performance is obtained than any other of the individual learning algorithms. In the field of Post-Silicon Validation for semiconductor devices (PSV), data sets are typically provided that consist of various devices like, e.g., chips of different manufacturing lines. In PSV, the task is to approximate the underlying function of the data with multiple learning algorithms, each trained on a device-specific subset, instead of improving the performance of arbitrary classifiers on the entire data set. Furthermore, the expectation is that an unknown number of subsets describe functions showing very different characteristics. Corresponding ensemble members, which are called outliers, can heavily influence the approximation. Our method aims to find a suitable approximation that is robust to outliers and represents the best or worst case in a way that will apply to as many types as possible. A 'soft-max' or 'soft-min' function is used in place of a maximum or minimum operator. A Neural Network (NN) is trained to learn this 'soft-function' in a two-stage process. First, we select a subset of ensemble members that is representative of the best or worst case. Second, we combine these members and define a weighting that uses the properties of the Local Outlier Factor (LOF) to increase the influence of non-outliers and to decrease outliers. The weighting ensures robustness to outliers and makes sure that approximations are suitable for most types. 
\end{abstract}


%
\IEEEpeerreviewmaketitle

\section{Introduction}
Ensemble learning is the process of generating and combining multiple, diverse models in order to solve particular Machine Learning tasks. Many applications of ensemble learning focus on classification problems. In this context, ensemble learning is also known as multiple classifier systems or ensemble systems. In addition to ensemble learning, ensemble systems cover the combination of different types of base models \cite{kn}. In the following, the term ensemble system is used to define the present task and to highlight the system of estimators, each trained on a specific subset of the data. In this application, the system is limited by the number of subsets. Additional subsets cannot be generated with sub-sampling strategies because each subset corresponds to a different entity.  \newline 
The primary objective of classification applications aims to improve the classification performance of the overall model. Note that regardless of the application, there is no guarantee that the combination of multiple models will always result in a better performance than the best individual model performs in the ensemble \cite{fg}. It is obvious that combining multiple models reduces the bias and variance of a learning algorithm \cite{hl, dtg}. This reduction effect is a primary reason to use ensemble learning. 
In \cite{dtg}, Dietterich lists three reasons in favour of using ensembles: (i) a statistical reason: the lack of adequate data results in an improper representation of the data distribution. Individual models suffer from high variance, but not their combination; (ii) a computational reason: different models may solve a given problem, but it is unclear which one to select. Combinations can exploit the strengths of multiple models; (iii) a representational reason: the model is not able to represent the data distribution. Individual models suffer from high bias but not combinations. \newline
In general, ensemble learning methods differ in two ways: (i) in the applied procedure to create individual models; (ii) in the strategy combining individual models. In the present task, the number of subsets are limited in (i), not in terms of different learning algorithms but by the number of subsets. In (ii), a strategy is developed having the objective to approximate underlying functions, e.g., best or worst-case approximations, instead of improving the accuracy of arbitrary ensembles. \newline 
Many frequently used ensemble learning methods, e.g., Boosting \cite{bp}, Bagging \cite{bl}, or AdaBoost \cite{fy} fuse individual models in the combination process and use strategies that rely on (static) algebraic combination rules, e.g., majority voting, maximum, minimum, sum, etc., that are non-trainable \cite{kl, pr}. 
Few methods, e.g., Stacked Generalization \cite{wdh}, learn an additional model on top of the ensemble system and thus use trainable instead of non-trainable combiners. Which combination strategy has to use cannot be answered in general as it strongly depends on the specific problem. Thus, no unique best 
combiner exists that works well for all problems \cite{kl, acc}. \newline
Circuits and systems often show large process variations in the context of semiconductor test and manufacturing. A suitable configuration of specific variables and registers - so-called tuning knobs - needs to be computed to guarantee that the circuits stay within their specified limits and meet performance goals like robustness or power consumption. To optimize a configuration, the behavior of devices must be modeled. A possible approach in PSV is to learn the behavior of each device in a supervised regression task and aggregate device-specific models to describe the general device behavior. The aggregation process in PSV often does not have the objective to improve performance. Instead, the interest lies in a robust worst-case approximation that applies to as many devices as possible. Due to additional time constraints an aggregation method is required that is efficient and flexible at the same time. In this application, a robust approximation of the worst case is particularly challenging because outliers in the given set of devices are expected, e.g., defective devices. Thus, algebraic combination rules, e.g., maximum or minimum, are not suitable due to their sensibility to outliers. The same applies to other algebraic combination rules because they are not suitable for worst-case approximations. \newline
To tackle this challenge, we propose a method that uses a trainable combiner, e.g., a NN, to learn a robust combination rule that approximates the worst case over all devices in the presence of outliers. The combination rule, which we call 'soft-max' or 'soft-min', is learned in a two-stage process: (i) the selection of $k$ devices that are representative for the worst case; (ii) the robust combination of selected devices. The weighting in (ii) uses the properties of the LOF \cite{bmm} to ensure robustness to outliers and an approximation that represents the worse case across devices. Consequently, the subsequent optimization process has a better chance to find a universal tuning law.

\section{Related Work}
To the best of our knowledge, there are only a few approaches that use trainable combiners \cite{wdh, bl1, spwd, spwd1, pva}. The works \cite{bl1, spwd, spwd1} are variants of the Stacked Generalization method in \cite{wdh}, which was invented for classification tasks. In \cite{bl1}, Breiman used Stacked Generalization in regression tasks. In \cite{spwd, spwd1}, Smyth and Wolpert deployed Stacked Generalization in unsupervised learning tasks, e.g., non-parametric multivariate density estimation.
Stacked Generalization or "stacking" is a technique in which the outputs of an ensemble of models are given as inputs to a second-level learning algorithm. This learning algorithm is trained to optimally combine the model outputs to obtain better final performance. Stacking has been applied successfully on a wide variety of problems, including spam filtering \cite{sg}, sensor design \cite{fl}, chemometrics \cite{xl}, and tasks on other, large collection of data sets, e.g., the Netflix Prize data set of the UCI Machine Learning repository \cite{tamj}. Recommendation systems, e.g. \cite{tamj,bbt}, often use additional meta-features in the stacking process. Moreover, it is common to use multiple levels of stacking. \newline
The approach that is most related to this work is \cite{pva}. Instead of using meta-features, the stacking algorithm in \cite{pva} is based on weighted nearest neighbors, which change the weightings assigned to the individual models depending on the distance between a particular instance and its neighbors. None of the shown approaches has task-specific limitations, e.g., in the size of the ensemble system. \newline 
Nonetheless, there are still open issues of stacking approaches \cite{ti}: (i) which type of model to choose as a high-level combiner?; (ii) which features to use as inputs to the high-level combiner?. To address these issues, we use a NN as a high-level combiner because they are powerful and act as general function approximators. Furthermore, the aggregation NN learns which features to use, and thus there is no need to define an explicit set of features.

\section{Study area and available data}
\subsection{Setup}
\begin{figure}[!t]
	\centering
	\includegraphics[width=0.5\textwidth]{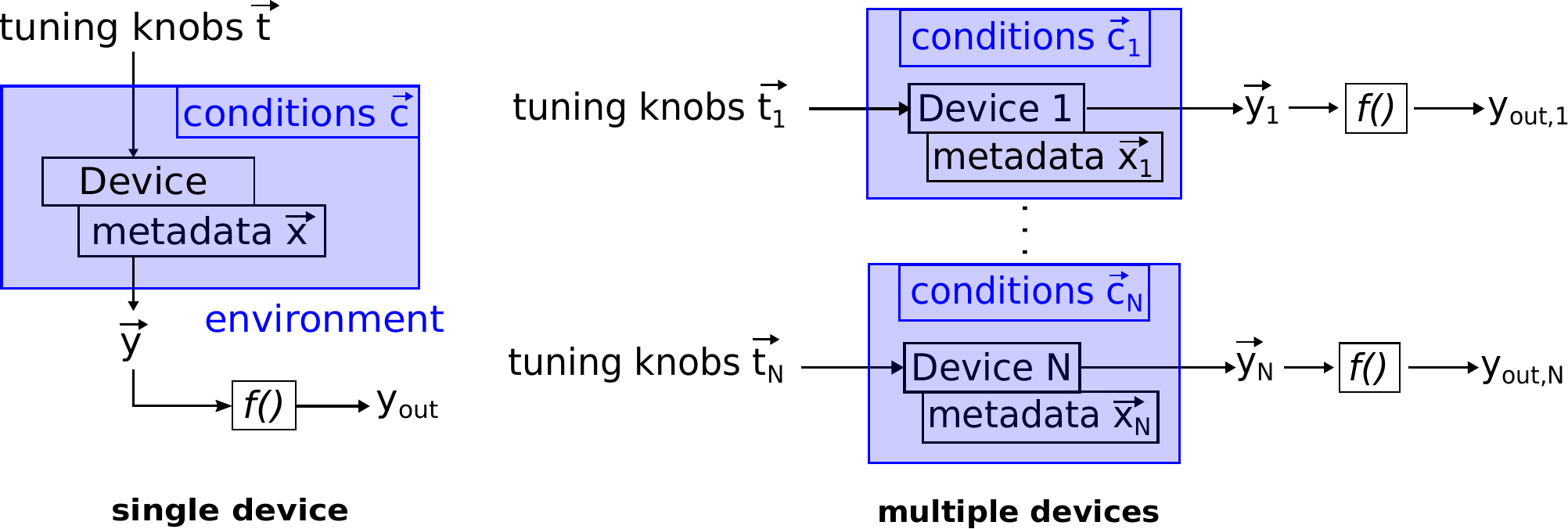}
	\caption{Experimental setup of the given data set. Inputs are tupels of tuning variables, (environmental) conditions, and metadata. Device i defines the mapping $(\vec{t_i}, \vec{c_i}, \vec{x_i}) \rightarrow \vec{y_i}$. Given $\vec{y_i}$, a single, scalar output value $y_{out, i}=f(\vec{y_i})$ is determined for each device. }
	\label{experimental_setup}
\end{figure}

In PSV, we aim for robust performance tuning to compensate impacts of process variations. These effects show as offsets or localized deformations of the output that appear at random and lead to abrupt changes of normal behavior. Due to millions of possible error locations, effects vary a lot, and there is no way to predict how outliers will behave. To determine a robust tuning, a set of $N$ devices is typically given. Thus, the ensemble system is limited to a total size of $N$ members. Each device has to be considered a black-box because the underlying functions of the devices are unknown. To analyze and model the device-specific behavior, devices are exposed to different environmental conditions $\vec{c}$, e.g., temperatures or voltages, having various tuning configurations $\vec{t}$. Additional metadata $\vec{x}$, e.g., information about the manufacturing process, is given for each device, see Fig. \ref{experimental_setup}. 
\subsection{Data sets} \label{ds}
\begin{figure}[!t]
	\centering
	\includegraphics[width=0.5\textwidth]{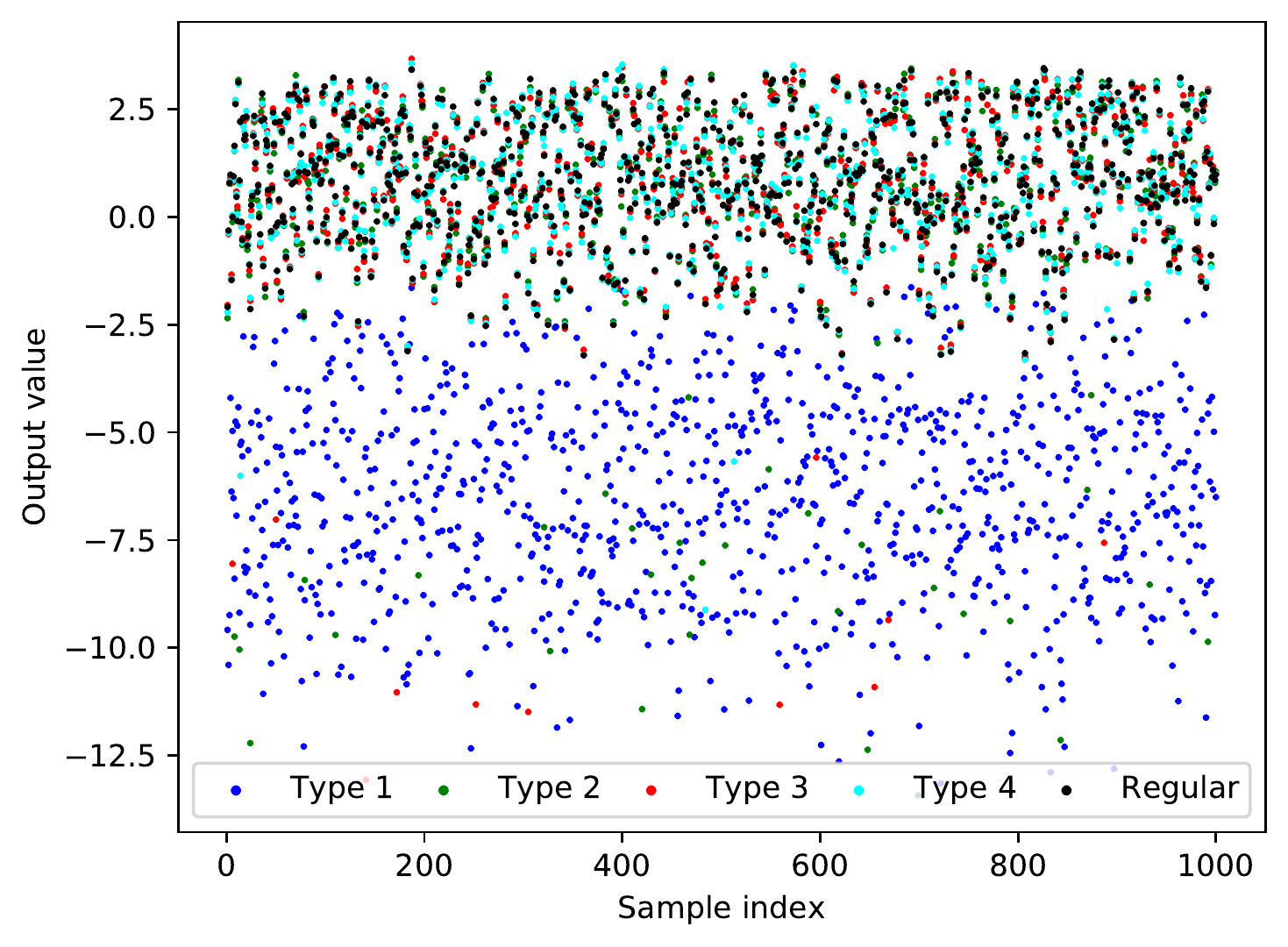}
	\caption{Output values of outliers (type 1 - 4) and a regular device. The plot shows output values $y_{out}$ for 1k randomly sampled input values.}
	\label{outlier_types}
\end{figure}

In the tuning stage of PSV, typically, 10-100k samples are generated per device. Thereby, input values should cover as much space as possible of the given parameter range. The total size of the data set is $N*D$, $D$ is the number of samples per device. \newline
In this research, two tuning data sets are used: (i) a real-world data set provided by Advantest. It consists of 9 devices with 100k samples per device. In general, inputs are of mixed data types (real number, categorical values, $\dots$), and the regression target is a real-valued scalar. Moreover, no prior knowledge of outliers is given, e.g., no information which devices are outliers; (ii) an artificial data set which we generate for a given number of devices and outliers, e.g., 30 devices with 4 outliers. To approximate the unknown mapping function of each device, we take the average output values that the devices in (i) produce. For non-outliers, we add a small noise value which we obtain by randomly sampling a normal distribution (mean $\mu=0.0$, standard deviation $\sigma=0.1$). We either add a constant offset or distort the output values of outliers in random areas of the input parameter space. In the latter case, we define the offset by a smooth offset function that is zero at the boundaries and maximum at the center of the randomly sampled area. To further add offsets with varying sign and amplitude, we include a probability $p_1$ and a random scaling factor $a$, e.g., $p_1=0.3$ and $a \in [-1, 1]$. With probability $p_1$, we choose the value of the smooth offset function, and with $1-{p_1}$, we choose zero. To scale the amplitude, we multiply the offset with $a$. Thus, the resulting artificial data set contains four different types of outliers with increasing difficulties to detect, see Fig. \ref{outlier_types}:
\begin{itemize}
	\item Type 1: Outliers with constant offsets
	\item Type 2: Outliers with an offset of smooth function
	\item Type 3: Outliers with an offset of smooth function and probability $p_1$
	\item Type 4: Outliers with an offset of smooth function, probability $p_1$, and scaling factor $a$
\end{itemize}

In the case of an artificial data set a prior knowledge of which devices are outliers is given. The artificial data set and the characteristics of different types of outliers are aligned with the expertise of Advantest to be as realistic as possible.

\subsection{Preprocessing}
We convert the different input data types to real values.  Due to different ranges of the input parameters, we normalize all input variables in the range -1 to 1 by applying a min-max normalization on the (real-valued) inputs, see \eqref{eq1}.
\begin{equation}
x'=-1 + \frac{2*(x-x_{min})}{x_{max}-x_{min}}\label{eq1}
\end{equation}
In case of device-specific modeling, we ignore metadata $\vec{x}$. Therefore, device-specific models learn the mapping $(\vec{t}, \vec{c}) \rightarrow y_{out}$.

\section{ORSA}
\begin{figure}[!t]
	\centering
	\includegraphics[width=0.5\textwidth]{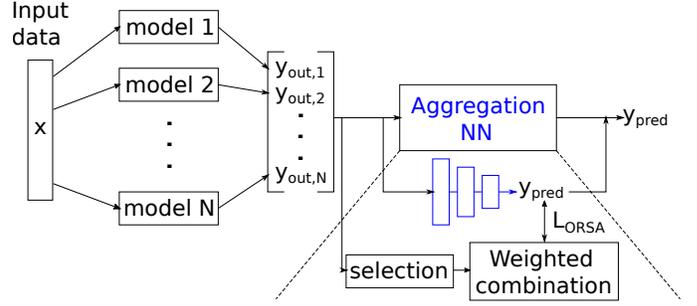}
	\caption{Illustration of the proposed method ORSA. It shows the selection and combination process as well as the calculation of $L_{ORSA}$, see \eqref{eq2}. The loss $L_{ORSA}$ is used to train the 'stacked' aggregation NN (highlighted in blue).}
	\label{orsa}
\end{figure}

Similar to most common ensemble system methods, our approach involves two key stages \cite{rza}: (i) a generation stage that generates individual ensemble models; (ii) a combination stage that aggregates the models to improve both final output and performance. This stage can involve a selection of individual ensemble members. In our case, it is impossible to generate an arbitrary amount of estimators because the number of devices limits the size of the overall ensemble system. Therefore, we train a single model for each device-specific subset of the data. The goal is to find a suitable combination strategy for the limited amount of models. Thereby, the models show differences in performance or accuracy due to different behaviors of the devices. Moreover, the models can not simply be made more accurate by an additional amount of effort in training. Altogether, we face a new task with different characteristics compared to ensemble learning.\newline
We focus on the model combination stage as there is still room for improvement and challenges \cite{acc}: (i) normalization issues that arise due to incomparable output scales. This may cause problems because individual members might be inadvertently favored; (ii) issues finding a suitable function to combine output values. Common functions are maximum or (damped/pruned) averaging. \newline 
In the following, it is assumed that the generation stage is finished and the individual, trained ensemble members are available. Thereby, the members can be any model of the underlying function, e.g., a NN or SVM. We consider a single sample $\vec{s}$ in the device-specific setup, which means that $\vec{s}=(\vec{t}, \vec{c})$. Each ensemble model $i$ has learned a specific mapping $f_i: \vec{s} \rightarrow y_{out,i}$. The overall ensemble output is $\vec{y_{out}}=[y_{out,1}, \cdots, y_{out,N}]=[f_1(\vec{s}), \cdots, f_N(\vec{s})] \in \mathbb{R}^N$. Depending on whether we aim for a robust approximation of the best or worst output value, we select the k largest or smallest values of $y_{out}$ respectively. Thus, we get $\vec{y_{out}}^k=[y_{out}^1, \cdots, y_{out}^k] \in \mathbb{R}^k$. The loss function of ORSA, which is used to train the model stacked on top of the ensemble, is defined as follows: 
\begin{equation}
\begin{split}
L_{ORSA} & =\sum_{i=0}^{k}w_i*(y_{out}^i-y_{pred})^2 \\
& =\frac{1}{\sum_{j=0}^{k}LOF_j}\sum_{i=0}^{k}\frac{1}{LOF_i}*(y_{out}^i-y_{pred})^2
\end{split}
\label{eq2}
\end{equation}
The loss definition in \eqref{eq2} is a weighted least square formulation that approximates a robust solution by minimizing the sum of squared errors made between the prediction and the true output values of the selected devices in $\vec{y_{out}}^k$. Note that there is no ground truth for the approximation task. Thus, ORSA is an unsupervised method. Furthermore, the weights $w_i$ are calculated as the reciprocal of the LOF and are normalized such that $\sum_{i} w_i = 1$. \newline 
The definition of LOF involves the calculation of the $k$-distance $d_k(A)$ of point $A$,  the reachability distance $rd_k(A)$, and the local reachability density $lrd_k(A)$. Given the value of parameter $k$, which defines the number of neighbors LOF is considering, the $k$-distance of any point is its distance to the $k^{th}$ nearest neighbor. With the $k$-distance, we can calculate the reachability distance as the maximum distance of two points, e.g., A and B, and the $k$-distance of the second point, see \eqref{eq3}.
\begin{equation}
\begin{split}
rd_k(A,B)= \max(d_k(B), d(A,B))
\end{split}
\label{eq3}
\end{equation}
The reachability distance $rd_k$ is used to calculate the local reachability density $lrd_k$ of point A. To get $lrd_k(A)$, we first calculate the reachability distance $rd_k$ to all k nearest neighbors and take their average. To get a density, we finally take the inverse, see \eqref{eq4}. The points that lie in or on the circle with radius $k$-distance and with center at point A are called $k$-neighbors and are denoted by $N_k(A)$.
\begin{equation}
\begin{split}
lrd_k(A) = \frac{1}{\left(\frac{\sum_{B \in N_k(A)} rd_k(A,B)}{|N_k(A)|}\right)}
\end{split}
\label{eq4}
\end{equation}
The LOF calculates $k$ ratios of $lrd_k$ of each point to its neighbors and take the average of these ratios, see \eqref{eq5}.
\begin{equation}
\begin{split}
LOF_k(A) = \frac{\sum_{B \in N_k(A)}\frac{lrd_k(B)}{lrd_k(A)}}{|N_k(A)|}
\end{split}
\label{eq5}
\end{equation}
In general, if a point is an outlier, its density should be smaller than the average density of its neighbors. Thus, the resulting LOF of an outlier is larger than 1. Non-outliers have comparable densities to their neighbors, and therefore the LOF is approximately 1. By using the (scaled) reciprocal of the LOF, our weighting ensures that points in low-density areas contribute less to the total loss in \eqref{eq2}. In other words, in contrast to the equally weighted case where $w_i=\frac{1}{k}$, $w_i<\frac{1}{k}$ in case of outliers. Similarly, the influence of non-outliers is increasing. \newline
With the definition in \eqref{eq2} and given value for the parameter $k$, we train a Feedforward NN on top of the individual ensemble members that learns a combination rule by minimizing the loss $L_{ORSA}$, see Fig. \ref{orsa}. Thus, our method allows a dynamic and outlier robust approximation of the best or worst case without any prior knowledge of outliers.

\section{Experiments and Results}
\subsection{Artificial data set} \label{ads}
In the following experiments, we use an artificial data set that consists of 30 devices. For each device, we generate 10k samples. Moreover, 4 randomly chosen devices are generated as outliers, one for each outlier type described in Sec. \ref{ds}. Here, the smooth offset function is a truncated normal distribution that is shifted and scaled in such that it is 0 at the boundaries $b_l$ and $b_u$ and -1 at the center $c=b_l + (b_u - b_l)/2$ in a randomly-chosen area of the parameter space. \newline
On top of the individual members, we stack an additional Feedforward NN. It consists of two (hidden) layers, the first with 64 and the second layer with 32 nodes. Finally, we use the data of all devices jointly to train the stacked NN in an unsupervised learning task. We do not split the data into a training and a validation set and we train the stacked NN for 25k training steps. Each training step updates the stacked NN for one batch, e.g., 64 samples, of the artificial data set. The qualitative analysis of the results focuses on three properties of our method: (i) the frequency with which ensemble members are selected for the set of $k$ worst devices; (ii) the loss contribution of each member; (iii) the outlier-sensitive weighting. At the end of this section, we discuss the properties of the parameter $k$ and how to choose suitable values for $k$ in the selection process and the calculation of the LOF.
\begin{figure*}[!t]
	\includegraphics[width=0.32\textwidth]{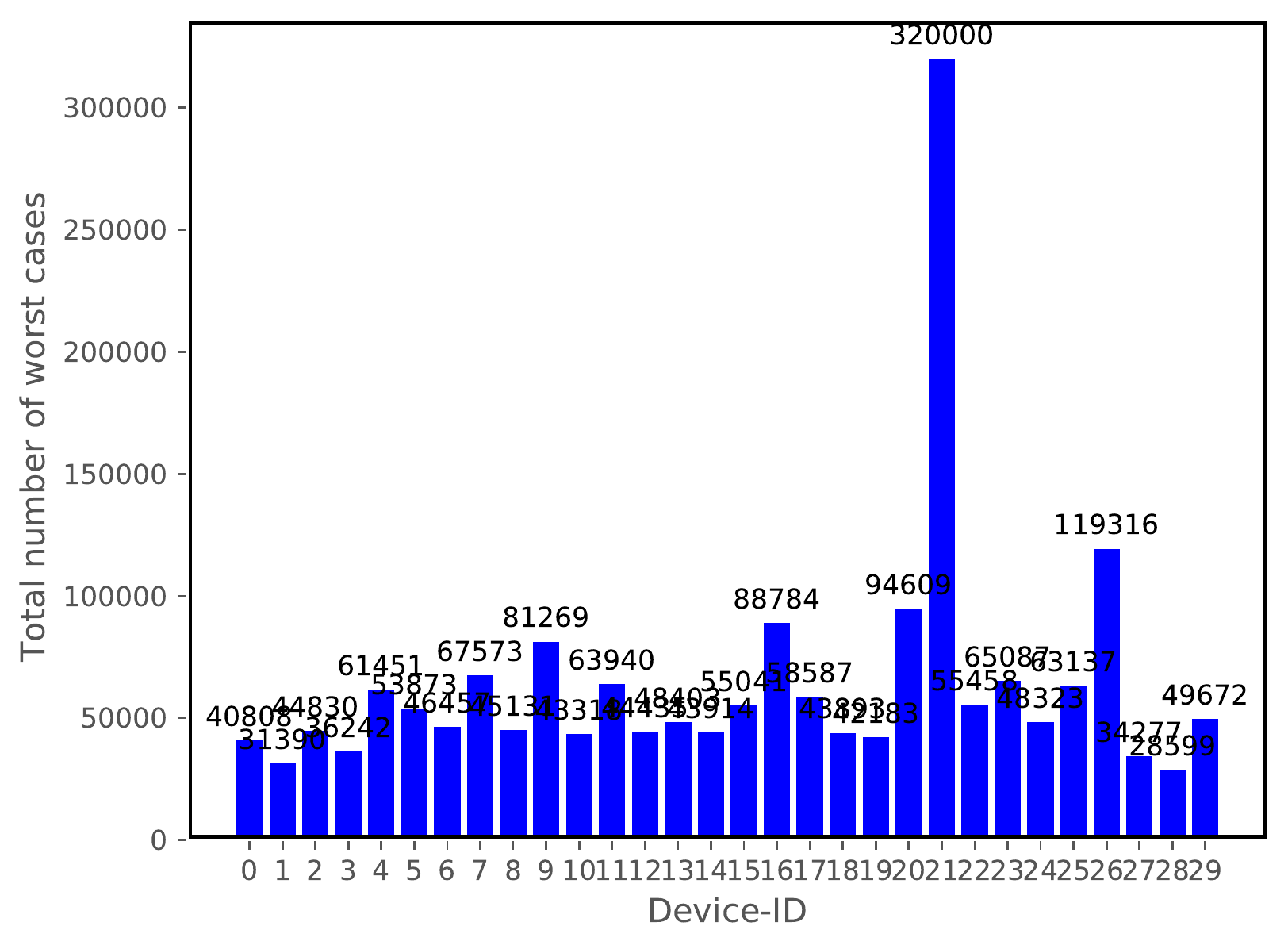} \hfill
	\includegraphics[width=0.32\textwidth]{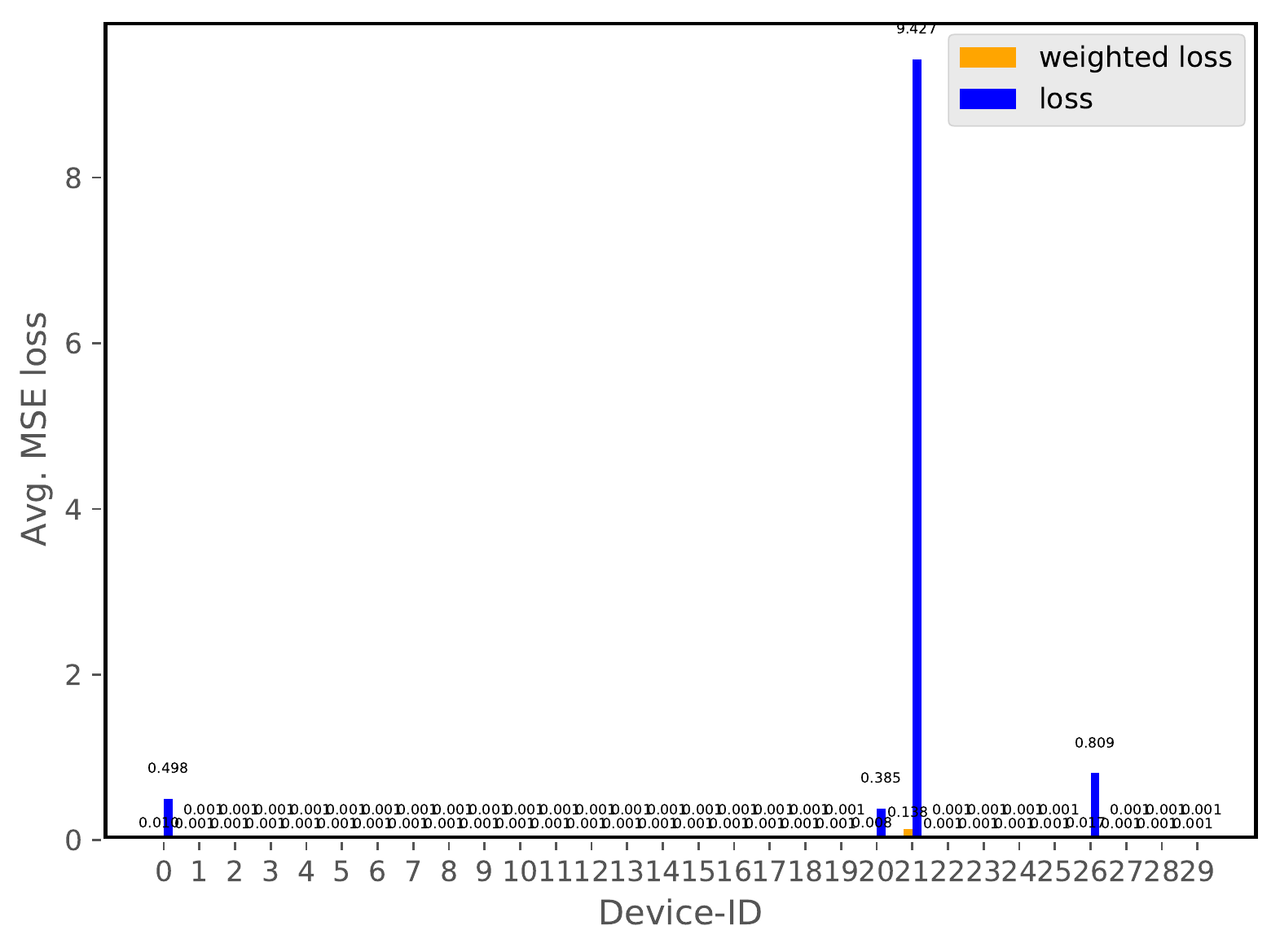} \hfill
	\includegraphics[width=0.32\textwidth]{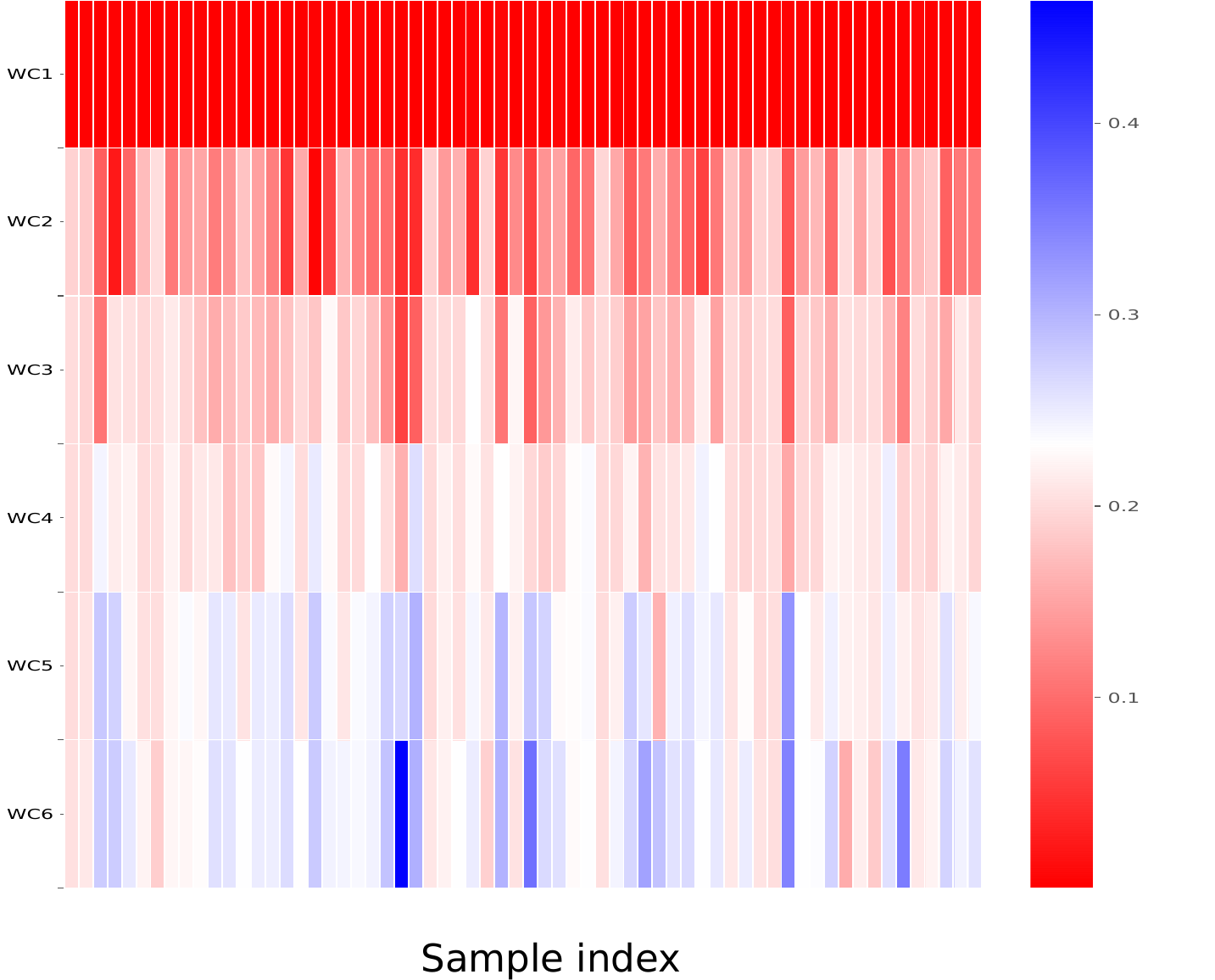}
	\caption{Final results for the artificial data set. The two bar plots show the results for the last 5k training steps. Thereby, a training step updates the NN for one batch, here 64 samples, of the data set. In addition, we compare the outlier weighted loss with the equally weighted case in the middle bar plot. The heatmap visualization on the right shows the outlier-sensitive weighting for a random batch of the data set.}
	\label{results_artificial_ds}
\end{figure*}
The results in Fig. \ref{results_artificial_ds} show the properties (i)-(iii) for $k=6$. In this configuration, we choose the same $k$ for the selection process and the calculation of the outlier-sensitive weighting, respectively. Moreover, we know that device 21 is a type 1, device 26 is a type 2, device 20 is a type 3, and device 0 is a type 4 outlier. Starting with device 21, we analyze the properties (i)-(iii) for the different outlier types.  \newline
Type 1: (i) due to the large, constant offset, outliers of type 1 are (almost) always in the set of $k$ worst devices; (ii) consequently, we observe that type 1 outliers have a large contribution to the total loss in \eqref{eq2}, especially in case of equal weights. Minimizing the equally weighted loss will result in a correction of the (worst-case) approximation towards device 21. We see that our outlier-sensitive weighting reduces the loss contribution of device 21 by a factor of almost 70 to prevent correction of faulty devices; (iii) the heatmap visualization confirms that weights of type 1 outliers are much smaller in comparison to the equally weighted case. Due to large offsets, the weights are reaching zero, which corresponds to our initial suggestion that the more a device is an outlier, the less their assigned weights. \newline
Type 2: (i) since we define the offsets of type 2 outliers by a smooth, Gaussian-shaped function that is non-zero only in a random area of the parameter space, the influence of type 2 is more subtle than of type 1. Still, we can observe that device 26 has the second-largest amount of worst cases; (ii) we expect the total loss contribution to be smaller than that of a type 1 outlier. Even though type 2 outliers are challenging to identify in the first bar plot, the second bar plot shows that our methods identify device 26 as the second most severe outlier. The outlier-sensitive weighting reduces the loss of device 26 by a factor of almost 5; (iii) the heatmap shows that our method does not reduce the weights of type 2 outlier to the same extent for all data points. In the random area of disturbance, we observe that our method reduces the weights almost to the same extent as of type 1 outliers. Outside this area, we see larger weights because, in these regions, outliers of type 2 behave like regular devices. Thus, our method considers device 26 to be more trustworthy. \newline
Type 3: (i) since we define the offsets of type 3 outliers as probabilistic, the effects of type 3 outliers are even more subtle than those of type 2. Probabilistic signifies that we take the value of the offset function with probability $p_1$, and with $p_2=1-p_1$, we set the offset to zero. Thus, it is hard to identify type 3 by observing the amount with which our method is selecting it. Although device 20 has the third most amount of worst cases, the bar plot shows that some regular devices have a similar amount; (ii) as a consequence of the probabilistic offset definition, we expect type 3 outliers to show smaller contribution to the total loss than type 2 outliers. The second bar plot shows that device 20, has the smallest contribution of all outlier types. However, the outlier-sensitive weighting reduces the loss of device 20 by a factor of almost 5, similar to type 2; (iii) the heatmap shows the impact of the probabilistic offset definition on the weighting. Because larger offsets appear with probability $p_1$, e.g., $p_1=0.3$, only a certain percentage of the weights, e.g., 30\%,  are reduced to the same extent as weights of type 2 outliers. Since type 3 outliers behave similar to regular types for larger regions of the parameter space,  the amount of non-reduced weights increases compared to type 2.\newline
Type 4: (i) as we scale the probabilistic offsets of type 4 outliers to random amplitudes, we expect  significantly different behavior only in small regions of the parameter space. Thus, identifying type 4 outliers is impossible in the first bar plot because device 0 has fewer worst cases than many regular devices; (ii) the second plot shows that our method considers type 4 as an outlier and thus reduces its loss contribution by a factor of almost 50. The total loss contribution of device 0 mainly depends on the amount of large or small amplitudes that we choose randomly; (iii) similar to type 3 outliers, larger offsets only appear with probability $p_1$. Due to the additional random scaling, we expect fewer cases of larger offset values. The heatmap shows that in comparison to type 3 our method increases weights of type 4 outliers. \newline
For all regular types, two aspects are valid: Firstly, the second bar plot in Fig. \ref{results_artificial_ds} shows that our method successfully minimizes the loss contributions of regular types to a value close to 0. Secondly, the corresponding weights of regular devices are larger than that of outliers, which means that our method puts a big effort into minimizing the loss contributions of the selected non-outliers as they are more trustworthy regarding the whole parameter space.
\subsection{Real-world data set} \label{rwds}
\begin{figure*}[!t]
	\includegraphics[width=0.32\textwidth]{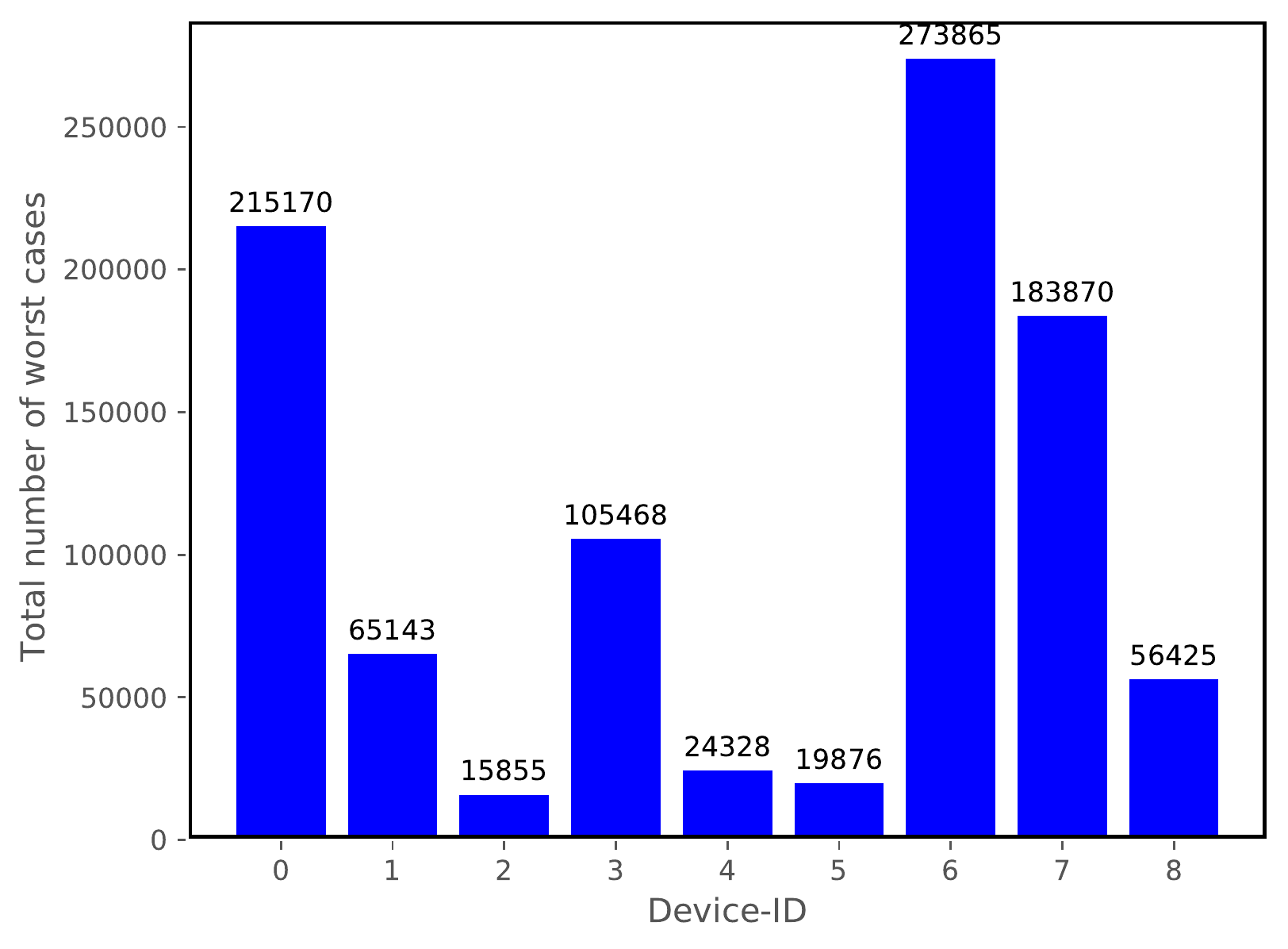} \hfill
	\includegraphics[width=0.32\textwidth]{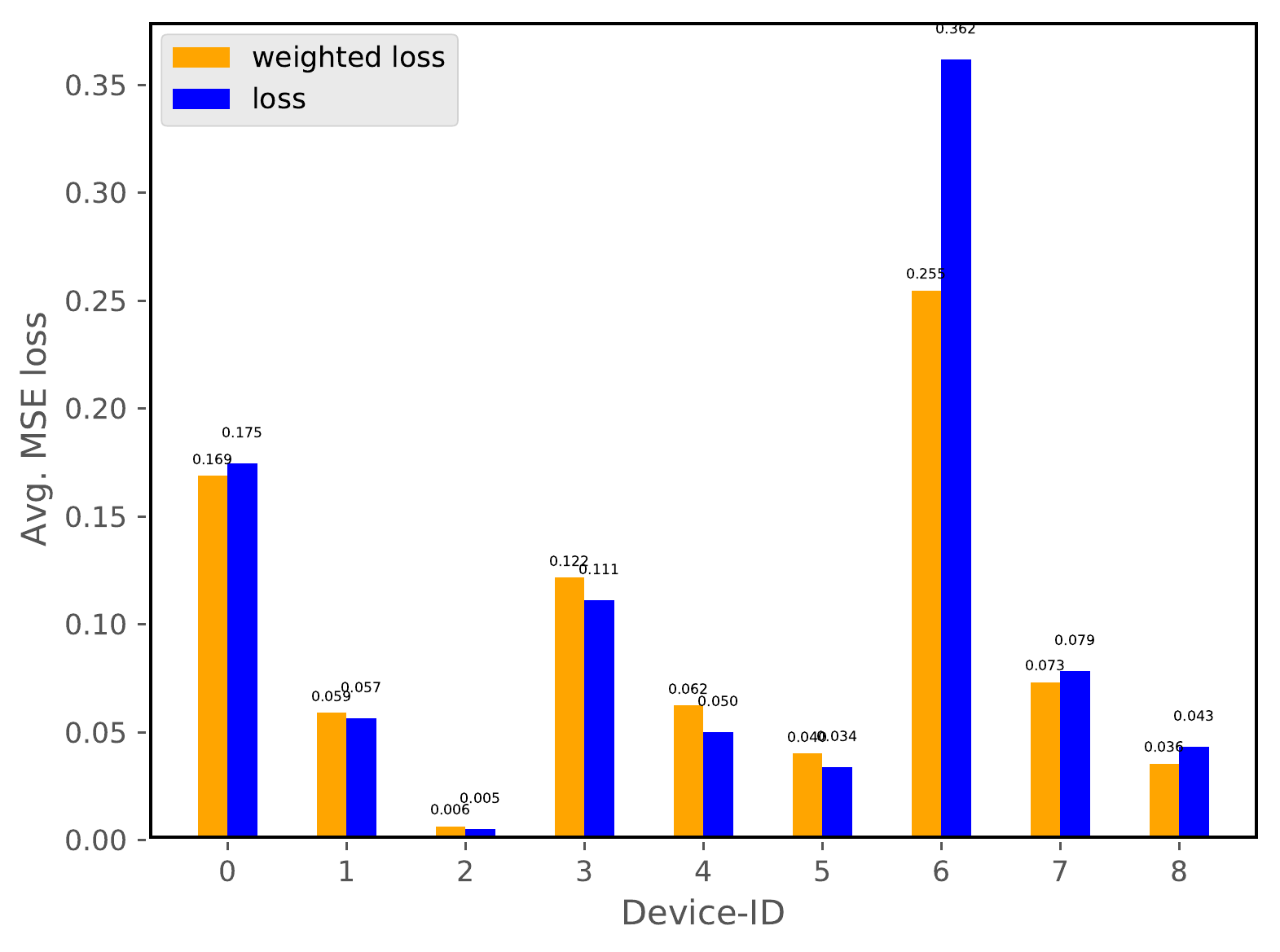} \hfill
	\includegraphics[width=0.32\textwidth]{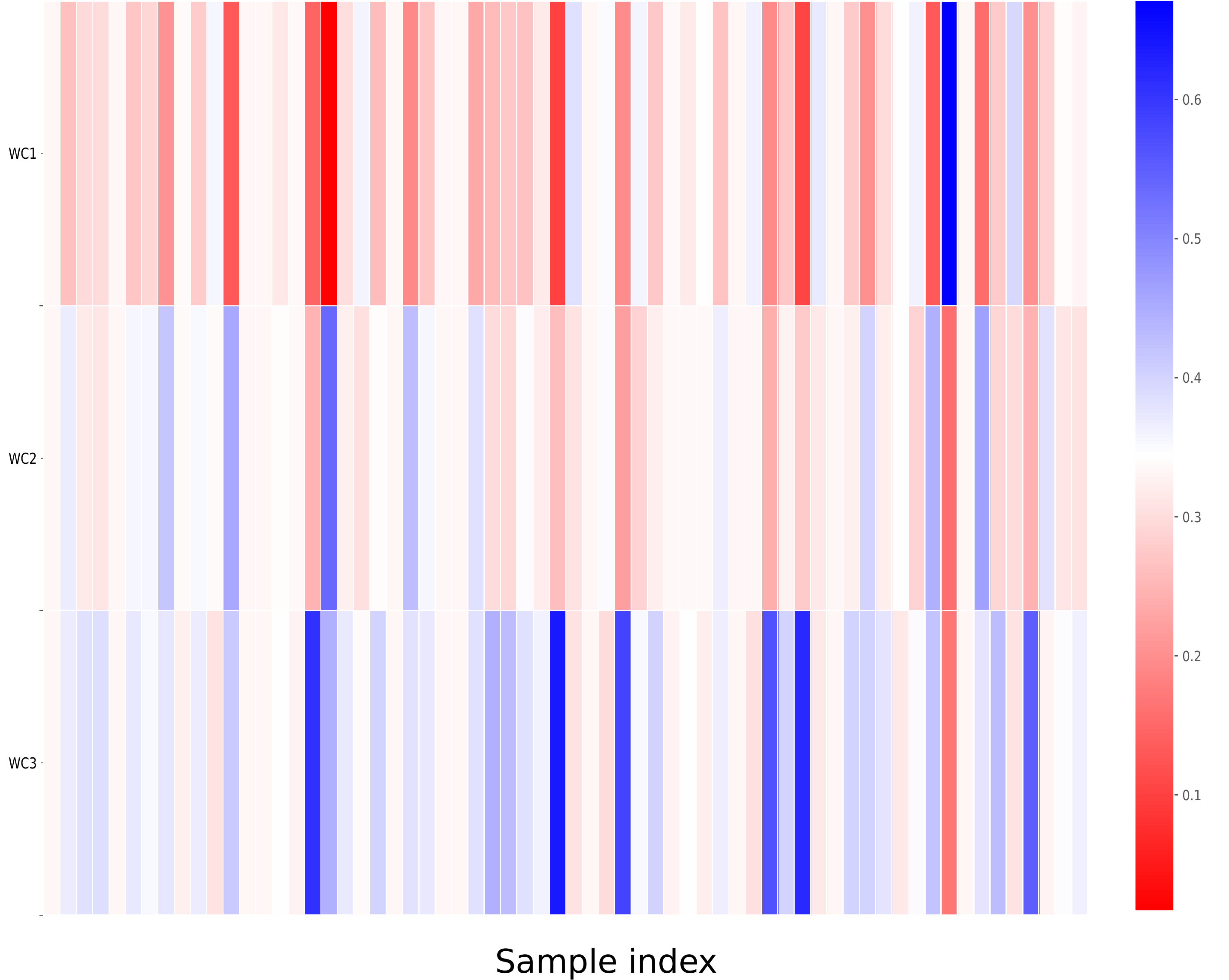}
	\caption{Final results for the real-world data set. The two bar plots show the results for the last 5k training steps. Thereby, a training step updates the NN for one batch, here 64 samples, of the data set. In addition, we compare the outlier weighted loss with the equally weighted case in the middle bar plot. The heatmap visualization on the right shows the outlier-sensitive weighting for a random batch of the data set.}
	\label{results_real_world_ds}
\end{figure*}

The experimental setup for the real-world data set is similar to the setup in \ref{ads}. The real-world data set consists of 9 devices with 100k samples per device. Moreover, we have no prior knowledge of the data including outliers and their types. We use the same architecture for the additional aggregation NN, and train it for 25k training steps, with a batch size of 64 samples. The results  are shown in Fig. \ref{results_real_world_ds}. For the real-world data set, we use $k=3$ in the selection process and the outlier-sensitive weighting. In the following, we qualitatively evaluate the results. Finally, we compare the final results of both experiments to analyze possible outlier types in the real-world data. \newline
The first plot in Fig.\ref{results_real_world_ds} shows that the real-world data set does not contain devices that are always in the set of $k$ worst devices. We observe that device 6 has a significantly higher amount of worst cases than the rest of devices. Regarding the total amount of worst cases, two other devices, namely devices 0 and 7, are noteworthy. In the first plot, it remains unclear whether devices 0, 6, and 7 are outliers or examples with poor performances. \newline
The second plot shows that our method reduces the loss contributions of devices 0, 6, 7, and 8, whereas it is slightly increasing the influence of remaining devices. In particular, device 6 has the highest contribution to the total loss and our method significantly reduces that contribution. In comparison to the results of the artificial data set, the significant reduction may be indicating that device 6 is a type 1 outlier, however, having a smaller offset value since the effect is less visible in the real-world data. For the remaining devices, the impact of increasing or reducing the loss contribution is not significant enough to infer if they are outliers and of which type they are. A reason for smaller impacts may be that the distortions of type 2 to 4 are even more subtle in the real-world data, e.g., in smaller areas of distortion. \newline
The heatmap in Fig. \ref{results_real_world_ds} shows that our method mainly reduces the weights of device 6. In contrast to the results in \ref{ads}, it seems that in the real-world data, type 1 outliers have smaller offsets that are not constant, e.g., function of a subset of input variables. By going from top to bottom, we observe that our method assigns increasing weights to the devices in the corresponding worst-case because our method considers them to be more trustworthy. Nonetheless, devices in the second-worst case, e.g. device 7, still show subtle effects that are similar to those of the outlier types 2 to 4 in the artificial data set. In general, we can conclude that our method detects unusual behavior of devices and successfully reduces their influence. Analogically, the influence of non-outliers, which are more similar to other devices, has been increased. Although assumptions about the type of outliers can be made, real-world data often include defective devices that show unusual behavior in various ways. In many cases, the effects of unusual behavior are very subtle and only appear in small areas of the possible parameter space. Thus, detecting and classifying defective devices without any prior knowledge, e.g., about different types of outliers, in a real-world data set remains a challenge.
\subsection{Setting of hyperparameter $k$}
\begin{figure*}[!t]
	\includegraphics[width=0.495\textwidth]{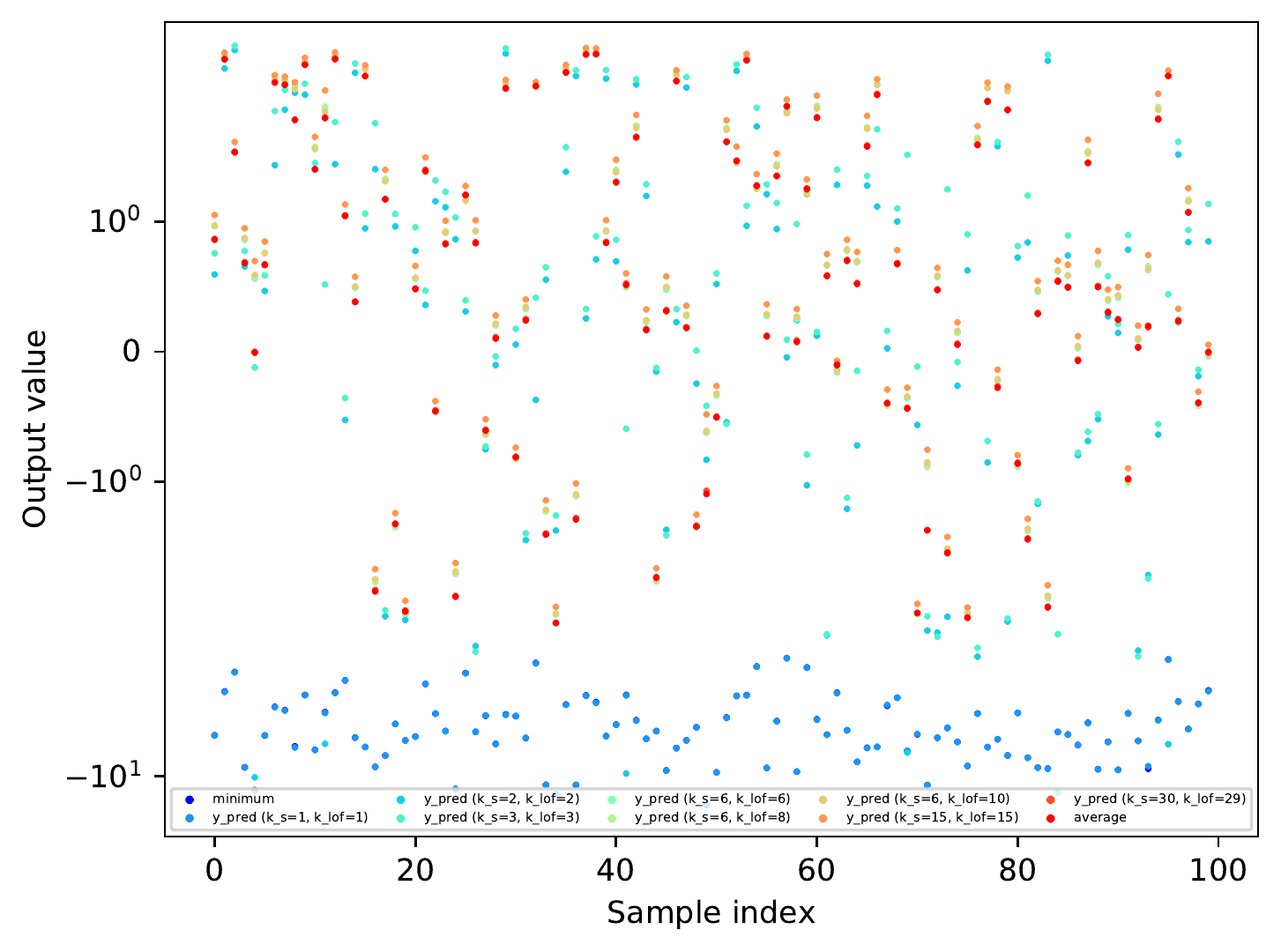} \hfill
	\includegraphics[width=0.495\textwidth]{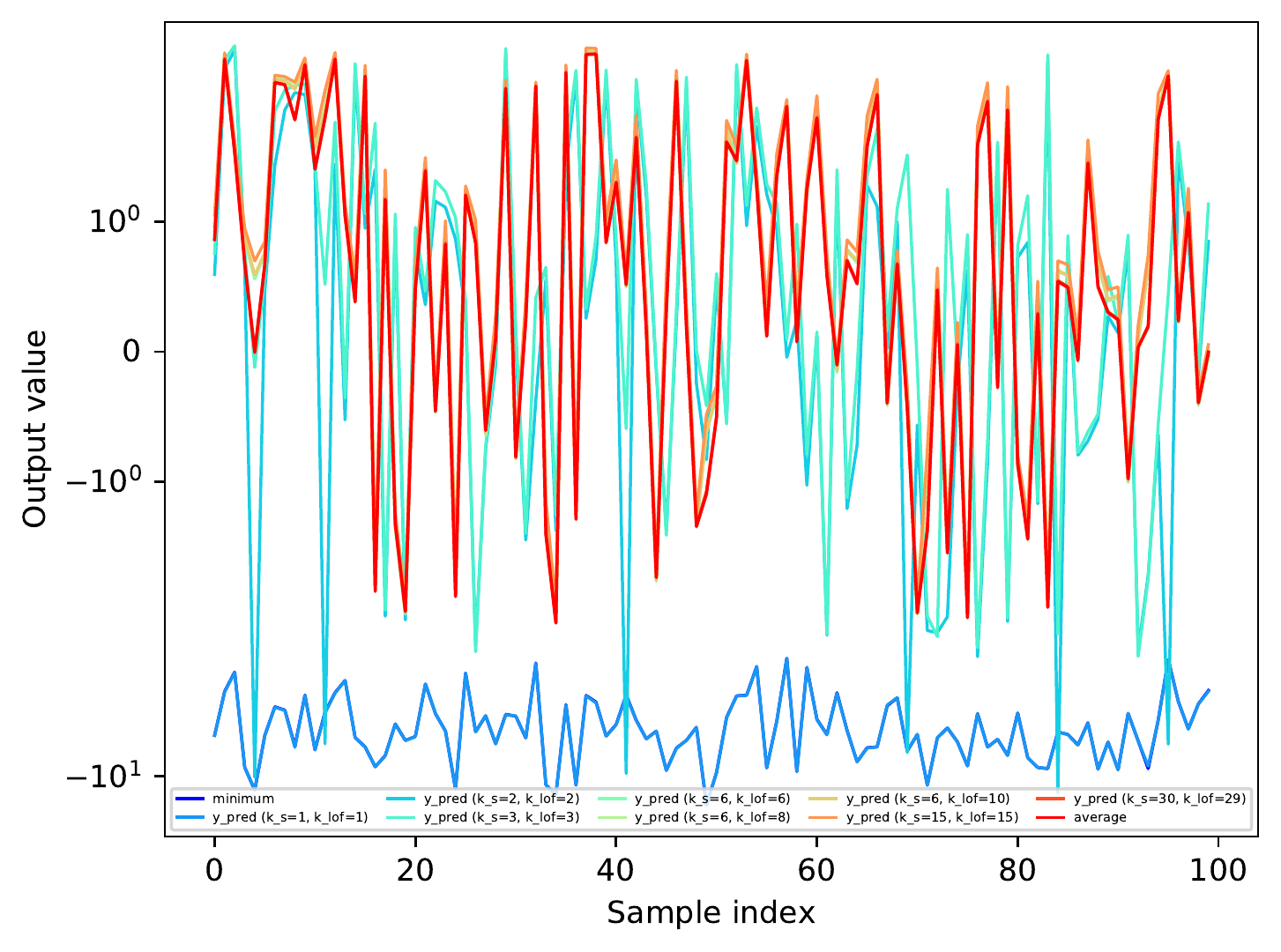} 
	\caption{Scatter and line plot of the output values of our method for 100 randomly chosen samples of the artificial data set. Outputs are shown for different hyperparameters settings. The predicted output values are compared to the minimum and the average of the (true) device outputs. In the minimal hyperparameter setting our method is equivalent to the minimum and for maximal setting to the average.}
	\label{setting_parameter_k_artificial_ds}
\end{figure*}

\begin{figure*}[!t]
	\includegraphics[width=0.495\textwidth]{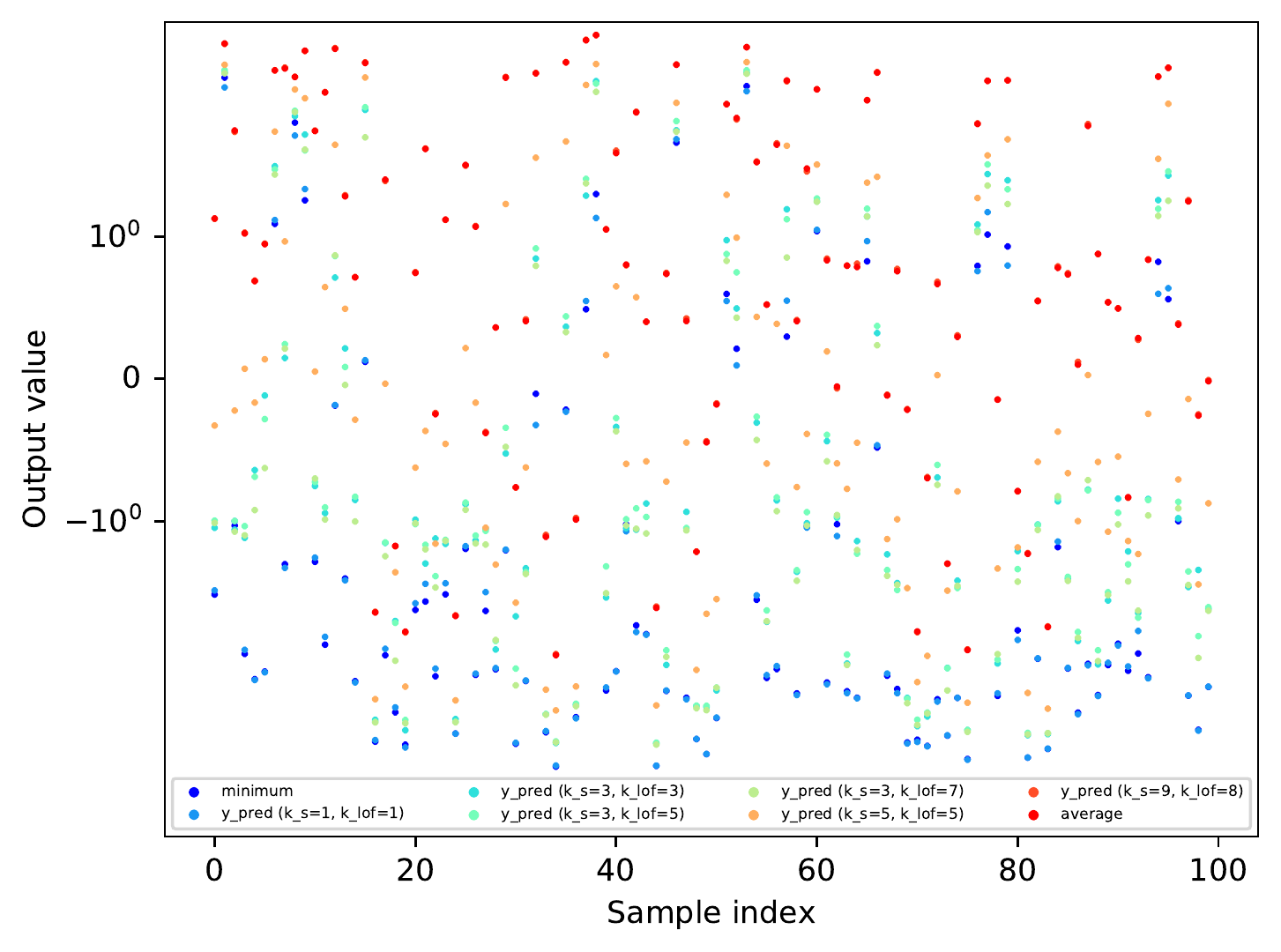} \hfill
	\includegraphics[width=0.495\textwidth]{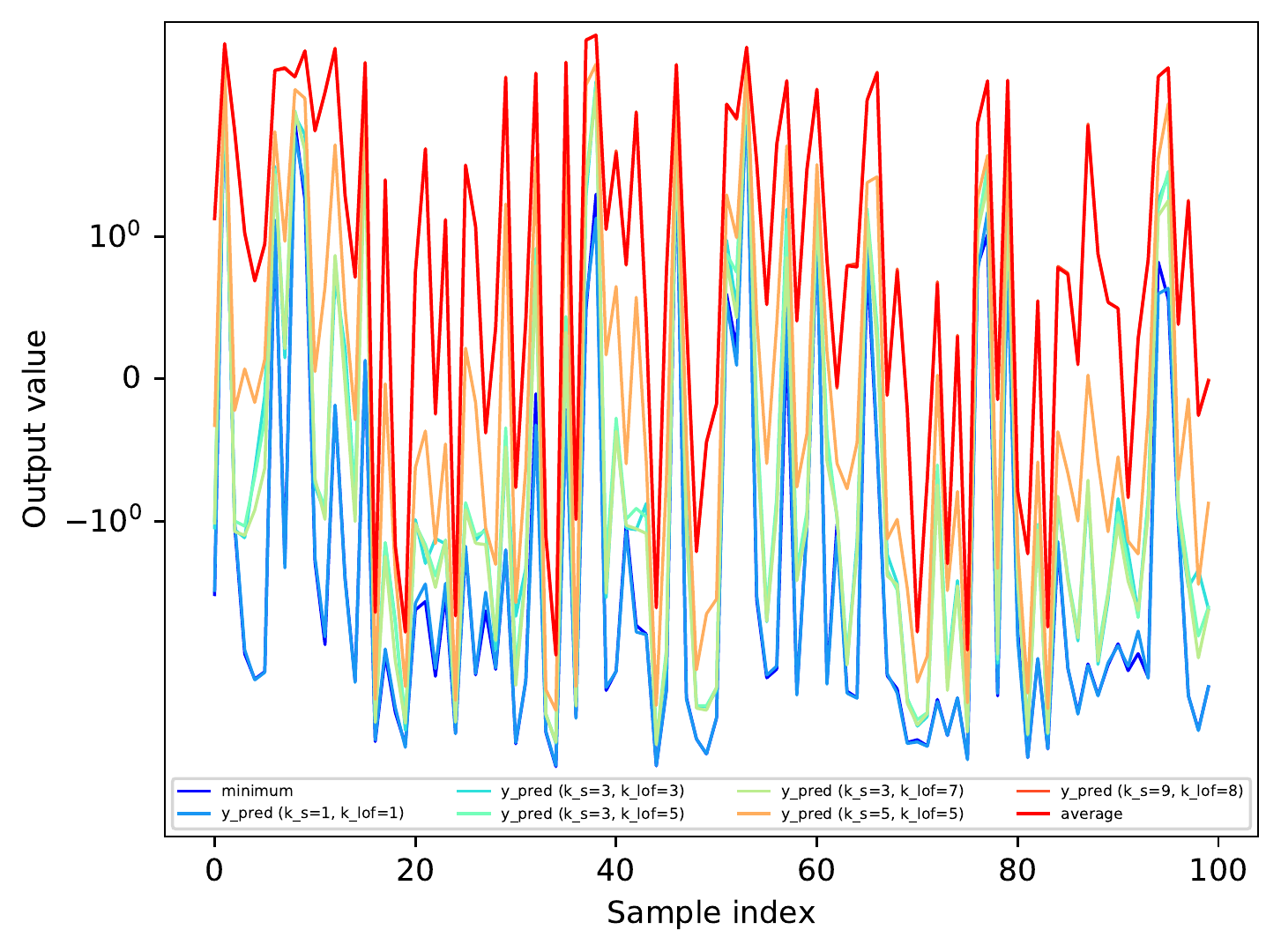} 
	\caption{Scatter and line plot of the output values of our method for 100 randomly chosen samples of the real-world data set. Output are shown for different hyperparameters settings. The predicted outputs values are compared to the minimum and the average of the (true) device outputs. In the minimal hyperparameter setting our method is equivalent to the minimum and for maximal setting to the average.}
	\label{setting_parameter_k_real_world_ds}
\end{figure*}

Section \ref{ads} and \ref{rwds} show the results of experiments in which we use the same value for $k$ in both the selection process and the calculation of the outlier-sensitive weighting. While we found this setup to work well in many experiments, in general, we have to distinguish between $k_s$ and $k_{lof}$. The parameter $k_s$ determines the number of devices we select for the worst case, and $k_{lof}$ determines the number of nearest neighbors we use in \eqref{eq3} - \eqref{eq5}. For the density-based definition of the weighting, a suitable rule of thumb is to choose $k_{lof} \geq k_s $. Due to increasing computational complexity for large $k_{lof}$, we usually select $k_s=k_{lof}$. Fig. \ref{setting_parameter_k_real_world_ds} and Fig. \ref{setting_parameter_k_artificial_ds} are comparing the influence of different values of $k_s$ and $k_{lof}$ in experiments realized with both the real-world data set and the artificial data set. For both data sets, we observe that our method predicts the hard minimum in case of minimal values for $k_s$ and $k_{lof}$, here $k_s=k_{lof}=1$. In the case of maximal values for $k_s$ and $k_{lof}$, we observe that our method learns the average function over all devices. Depending on the characteristics of the underlying data set, choosing values between the minimum and maximum value of both $k_s$ and $k_{lof}$, our method predicts outputs that are a trade-off between the hard minimum and the average of all device outputs. \newline
Fig. \ref{setting_parameter_k_artificial_ds} shows the results of different hyperparameter settings for the artificial data set. Because outlier devices behave very differently in comparison to any regular device and due to similar behavior among regular devices, we observe that our method is robust for a large number of possible hyperparameter values. Thus, only for very small values, here $k \le 3$, Fig. \ref{setting_parameter_k_artificial_ds} shows a significant difference between the output of our method and the average of all devices. \newline
Fig. \ref{setting_parameter_k_real_world_ds} visualizes the results of different hyperparameter settings for the real-world data set. Here, it is observable that the larger $k_s$ or rather $k_{lof}$ is, the more our method predicts output values closer to the average. \newline
In order to summarize, choosing $k_s=k_{lof}$ is a suitable default setting. Moreover, it is advisable to start with lower values and to incrementally increase them if the underlying data set appears to contain multiple outlier devices. Depending on a prior assumption getting a small or larger number of faulty devices, the value of the hyperparameter should be chosen accordingly.

\section{Conclusion}
In this paper, we proposed an outlier robust aggregation methodology. The method has been demonstrated for worst-case approximations of device outputs, a typical but challenging task in PSV. It has been shown, through experiments on both artificial and real-world data, that our method is able to detect outliers of different types and to reduce their influence on the (worst-case) approximation. Moreover, the qualitative analysis of the real-world data has revealed which devices behave differently and which type of outliers these devices are. On the one hand, the experiments have shown that the weighting is outlier-sensitive and suitable for the present task. On the other hand, individually calculating the density-based weighting for each sample, which includes the distance calculation to $k$ nearest neighbors, can be computationally complex. Furthermore, it is necessary to find suitable settings for the hyperparameter values. Nonetheless, the proposed method has shown encouraging results and may be well suited for a wider variety of tasks, in which worst or best case approximations are needed, e.g., sensor fusion or weather forecasts based on multiple predictions of different providers. Finally, the approximation properties for different hyperparameter settings are being discussed, and empirical parameter setting instructions are provided. \newline
In the future, the project will be to learn an outlier-sensitive weighting instead of individually calculating weights based on the LOF. Furthermore, clustering techniques, visualization analysis, and task-specific selection criteria may help to find suitable values for the hyperparameters. 


\section*{Acknowledgment}

This research was supported by Advantest as part of the Graduate School "Intelligent Methods for Test and Reliability" (GS-IMTR) at the University of Stuttgart.



\bibliographystyle{IEEEtran}
\bibliography{IEEEabrv,icmla21_literature}
\end{document}